\pdfoutput=1

\documentclass[11pt]{article}

\usepackage{EMNLP2023}

\usepackage{times}
\usepackage{latexsym}

\usepackage[T1]{fontenc}

\usepackage[utf8]{inputenc}

\usepackage{microtype}

\usepackage{inconsolata}

\usepackage{amsmath}
\usepackage{graphicx}
\usepackage{booktabs}
\usepackage{amssymb}
\usepackage{transparent}
\usepackage{multirow}
\usepackage{tikz}
\usepackage[capitalize]{cleveref}
\usepackage{pifont}
\usepackage[super]{nth}
\usepackage[shortlabels]{enumitem}
\usepackage{quoting}
\usepackage{subcaption}
\usepackage{adjustbox}
\usepackage[ruled,vlined]{algorithm2e}

\usepackage{array}
\usepackage{booktabs}
\usepackage{tablefootnote}

\newcommand\blfootnote[1]{%
  \begingroup
  \renewcommand\thefootnote{}\footnote{#1}%
  \addtocounter{footnote}{-1}%
  \endgroup
}
\usepackage[hang,flushmargin]{footmisc} 

\usepackage{fontawesome5}

\title{\textsc{Muted {\color{red} \faVolumeMute}}: Multilingual Targeted Offensive Speech Identification and Visualization}

\author{Christoph Tillmann, Aashka Trivedi, Sara Rosenthal, Santosh Borse \\ \bf{Rong Zhang}, Avirup Sil, Bishwaranjan Bhattacharjee \\
         IBM Research AI}

\begin{document}
\maketitle
\begin{abstract}
Offensive language such as hate, abuse, and profanity (HAP) occurs in various content on the web. While previous work has mostly dealt with sentence level annotations,
there have been a few recent attempts to identify \emph{offensive spans} as well. 
We build upon this work and introduce \textsc{Muted}, a system to identify multilingual HAP content by displaying offensive \textsc{arguments} and their \textsc{targets} using heat maps to indicate their intensity. 
\textsc{Muted} can leverage  any transformer-based HAP-classification model and its attention mechanism out-of-the-box to identify toxic spans, \emph{without further fine-tuning}.
In addition, we use the spaCy library to identify the specific \textsc{targets} and \textsc{arguments} for the words predicted by the attention heatmaps.  We present the model's performance on identifying offensive spans and their targets in existing datasets and present new annotations on German text. Finally, we demonstrate our proposed visualization tool on multilingual inputs. 



\blfootnote{\color{red} \textsc{WARNING:} This paper contains offensive examples.}
\end{abstract}

\section{Introduction}

\begin{figure}[t!]
\begin{subfigure}{0.48\textwidth}
\centering
\includegraphics[clip,width=0.8\textwidth]{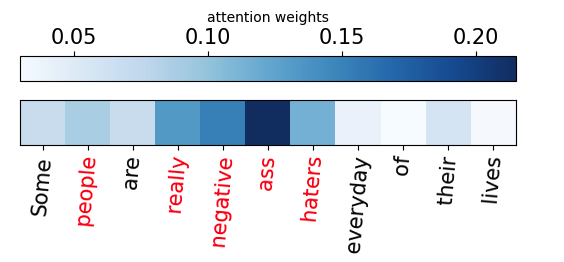}
\caption{Attention Heatmap}
\end{subfigure}
\bigskip
\begin{subfigure}{0.48\textwidth}
\centering
\includegraphics[clip,width=0.8\textwidth]{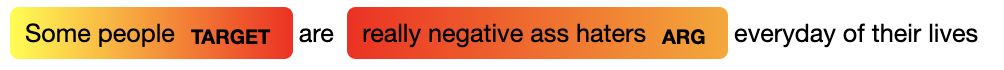}
\caption{SpaCy: visually identifying Target and Argument}
\end{subfigure}
\bigskip
\caption{Example system output that shows the intensity of the offensive \textsc{argument} and its \textsc{target},<\textsc{t,a}>: (a), (b): <people, really negative ass haters> .}
\label{fig-example-1}
\end{figure}

\begin{figure}
\begin{subfigure}{0.48\textwidth}
\centering
\includegraphics[width=0.8\textwidth]{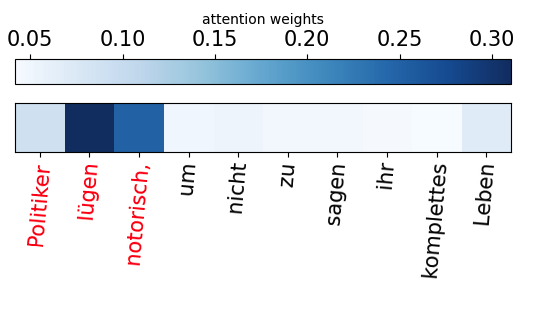}
\end{subfigure}
\caption{German Input (\textit{"Politicians notoriously lie, not to say their entire lives"}): <Politiker, lügen notorisch> .}
\label{fig-example-2}
\end{figure}

\begin{figure*}[t!]
\centering
\includegraphics[width=\textwidth]{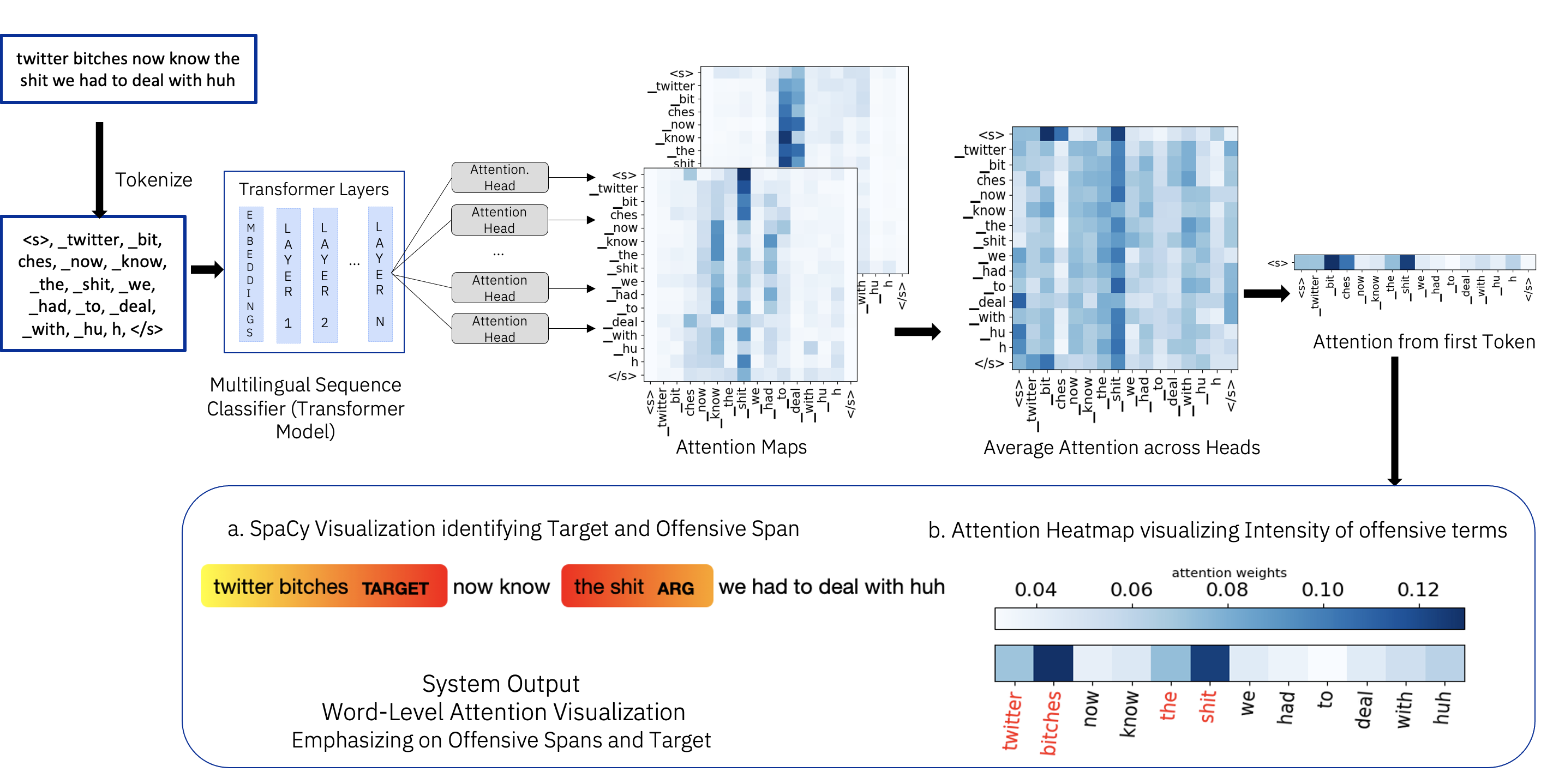}
\caption{\textsc{Muted} {\color{red} \faVolumeMute}: Visualizing offensive spans and targets using Attention Heatmaps. A token-level attention score of a given sentence is obtained using the average attention across all heads of the last layer of the given HAP classifier, and extracting the attention from the first token (often the CLS vector). The score for a word is calculated as the maximum token-level attention score of its constituent tokens. Finally, we display the predicted spans using the attention heatmap, and use spaCy's dependency parser to identify the target and argument in the predictions. }
\label{fig-flowchart}
\end{figure*}  

Offensive language such as hate, abuse, and profanity (HAP) occurs in various content on the web such as social media sites (e.g. Twitter) and discussion forums (e.g. Reddit). Such content can be hurtful to the reader, and identifying and visualizing HAP speech is necessary to understand and avoid harm. It increases interpretability and can be used to hide and provide a warning for offensive terms, and to avoid generating hate in large language models. 

While such visualizations exist, the focus has primarily been on English HAP and on identifying offensive language on the sentence level  \cite{mcmillan-major-etal-2022-interactive}. There are few works that explore spans and other languages \cite{ranasinghe-zampieri-2021-mudes, 10.1145/3449280} but these do not identify and visualize the \textsc{target} of the offensive \textsc{argument} which is an important indicator regarding whether the offensive argument is harmful or not, as shown in \citet{zampieri-etal-2023-target}.

We propose identifying hate using existing approaches \cite{caselli-etal-2021-hatebert} to display multilingual offensive \textsc{arguments} and their \textsc{targets} using heat maps as a means of showing their intensity. Moreover, the spaCy library \cite{spacy} can also be used to identify the specific target and argument from the predicted words. An example with a <\textsc{t,a}> pair is shown for English and German inputs in \cref{fig-example-1} and \cref{fig-example-2}, with the resulting visualizations. Our contributions are as follows: 

\begin{itemize}[topsep=0pt, itemsep=0pt, leftmargin=.125in, parsep=0pt]
    \item We present \textsc{Muted}: A \textbf{MU}ltilingual \textbf{T}argeted \textbf{D}emonstration providing an intuitive way of visualizing existing classifiers by using transformer attention to identify the target of the offensive text as well as the offensive span.
    \item Unlike similar token classification techniques \cite{ranasinghe-zampieri-2021-mudes}, our system can be used with off-the-shelf hate/abuse/profanity detectors.
    \item Our approach is multilingual and we demonstrate it on English \cite{zampieri-etal-2023-target} and a new German targeted offensive speech dataset. In the future, we plan to extend to more languages, e.g. a Spanish data set.
    \item We present easy-to-use Python notebooks and a front-end UI to run our approach on any encoder-only HAP classifier to visualize the offensive <\textsc{t,a}> pair using heat-maps and spaCy \footnote{https://spacy.io/api/dependencyparser}. 
\end{itemize}


\noindent The rest of this paper describes related work, our approach for detecting offensive speech, and our model which outperforms existing sentence classifiers on the TBO \cite{zampieri-etal-2023-target} and TSD \cite{pavlopoulos-etal-2021-semeval} datasets. Finally, we present our system demonstration and its efficiency.



\section{Related Work}

Identifying offensive content has been a popular area of research in recent years \cite{Davidson_Warmsley_Macy_Weber_2017,JAHAN2023126232}. One popular model that is available is HateBERT \cite{caselli-etal-2021-hatebert} which is a Bert-based model finetuned on offensive speech from Reddit comments. Similar models exist in other languages such as deHateBERT\cite{aluru2020deep} in German. We present our own multilingual model for detecting offensive content which outperforms HateBERT on offensive span selection. However, our notebooks demonstrating our approach for identifying the offensive target and argument can be used with any transformer-based offensive classifier.

Several demos on offensive text exist that perform on the span or sentence level, mostly in English \cite{mcmillan-major-etal-2022-interactive, 10.1145/3449280}. Perhaps the most relevant demo is MUDES \cite{ranasinghe-zampieri-2021-mudes}. They identify offensive spans in input text by classifying each token as offensive or not, and support English, Danish and Greek. The UI is token-classification based, and can be used with their trained models and the datasets used in the paper (or any input text) to identify offensive spans which will be displayed in red. In contrast to other prior work, our heat map-based system can be used to visualize the offensive argument and target for any language for which a sentence level hate classifier is available.

\section{Approach}\label{sec:approach}


\textsc{Muted} provides an intuitive visualization of existing HAP classifiers by using attention maps to identify offensive text and their targets, as shown in \cref{fig-flowchart}. Formally, for a transformer model (of $L$ transformer layers and $H$ attention heads) finetuned to classify whether a given input sentence $x$ contains offensive language, we first obtain the attention outputs $A^L_i \in \mathbb{R}^{|x| \times |x|}, i\in [1, H]$ of the last transformer layer. We then compute the average attention across all heads, $A' = \dfrac{1}{H}\sum_{i=1}^{H}A^L_i $, and extract the attention vector for the first token (e.g., the CLS token for BERT \cite{devlin2019bert} models), $A'_0 \in \mathbb{R}^{1 \times |x|}$. Based on a threshold, we obtain the set of tokens $T$ with the highest attention score, which can be intuitively viewed as the tokens that contribute most to the classification decision. We convert the token-level attentions into word-level attentions by assigning a word the maximum attention of any of its constituent tokens. We provide the word-level attention visualization in the form of heat maps, and mark the target and the argument of the offensive span in the sentence (see the System Output in \cref{fig-flowchart}). 

Our system can be used to visualize any transformer-based model that is trained to classify if a given sequence has HAP content or not, irrespective of the language. In this work, we present the \emph{Piccolo-HAP classifier}\footnote{https://medium.com/@alex.lang/fair-is-fast-and-fast-is-fair-ibm-slate-foundation-models-for-nlp-3508412a4b04}, a tiny 4-layer XLM-Roberta \cite{conneau2020unsupervised} model (with 153 Million parameters) finetuned on the HAP detection task for 6 languages (English, German, Japanese, Spanish, French and Portuguese). Specifically, we distil the self-attention relations of an in-house XLM-Roberta Base Model on a task-agnostic (general purpose) manner into a 4-layer architecture, as proposed in \citet{wang2021minilmv2}. We finetune this general purpose language model on the HAP classification objective, using open-source multilingual annotated datasets \cite{founta2018large, Davidson_Warmsley_Macy_Weber_2017, rottger-etal-2021-hatecheck, gibert2018hate, ousidhoum-etal-multilingual-hate-speech-2019, jigsaw, haternet,germeval,german-reliability-hate-speech-annotations,leite2020toxic} originating from social media data, as well as internally annotated samples from CC100 \cite{conneau2020unsupervised} and scraped news data from the internet in the six languages mentioned above. For non-English data, we also translate English datasets \cite{Davidson_Warmsley_Macy_Weber_2017, founta2018large} to the language required.  We finetune the model on a total of 1.7 million sentences, with the majority of data being in English.

\section{Experiments}

We compare our model to a random baseline, as well as open-source toxicity classifiers (monolingual and multilingual). First, we evaluate a \textbf{random} selection of spans as target and arguments in the sentence. Specifically, each span in the sentence is marked as HAP with a probability of 0.50.  We also use three off-the-shelf English \textbf{HateBERT} models \cite{caselli-etal-2021-hatebert}, each finetuned on either Hateval \cite{basile-etal-2019-semeval}, Offenseval \cite{zampieri-etal-2019-semeval} or Abuseval \cite{caselli-etal-2020-feel}. These models were made available by the HateBERT authors\footnote{Model Repository for HateBERT: https://osf.io/tbd58/}, and we have not finetuned them ourselves. We also compare our multilingual model to another open-source multilingual classifier available on HuggingFace, \textbf{Multilingual Toxicity Classifier Plus [MTC+]}\footnote{\scriptsize{EIStakovskii/xlm\_roberta\_base\_multilingual\_toxicity\_classifier\_plus}}, and two German (monolingual) classifiers, \textbf{DeHateBERT-de}\footnote{Hate-speech-CNERG/dehatebert-mono-german}\cite{aluru2020deep}
and \textbf{German Toxicity Classifier Plus (V2)}\footnote{EIStakovskii/german\_toxicity\_classifier\_plus\_v2}.



\begin{table*}[t!]
\centering
\begin{adjustbox}{max width=\textwidth}
\begin{tabular}{|c|c|ccc|}
\hline
\multirow{2}{*}{\textbf{Model}} & {\textbf{TSD: F1 Score $\uparrow$}} &\multicolumn{3}{c|}{\textbf{English TBO: F1 Score $\uparrow$}} \\
& \textsc{Target Only} & \textsc{Target + Arg.} & \textsc{Arg. Only} & \textsc{Target Only}  \\
\hline
Random & 0.08&  0.19& 0.16& 0.13 \\
HateBERT (AbusEval) & 0.15 & 0.36 & 0.30 & 0.24  \\
HateBERT (HatEval) & 0.16 & 0.36 & 0.30& 0.27\\
HateBERT (OffenseEval) & 0.23& 0.43& 0.37& \textbf{0.34}\\
HF Multilingual Toxicity Classifier Plus & 0.29 & 0.36& 0.31 & 0.22 \\
Piccolo-HAP (Ours) & \textbf{0.51} & \textbf{0.50} & \textbf{0.50} & 0.32   \\
\hline
\end{tabular}
\end{adjustbox}
\caption{\label{tab:results-english}
Results on the TSD and TBO datasets (English). Best results in bold.}
\end{table*}


\begin{table*}[t!]
\centering
\begin{adjustbox}{max width=.85\textwidth}
\begin{tabular}{|c|ccc|}
\hline
\multirow{2}{*}{\textbf{Model}}  &\multicolumn{3}{c|}{\textbf{German TBO: F1 Score $\uparrow$}}  \\
 & \textsc{Target + Arg.} & \textsc{Arg. Only} & \textsc{Target Only} \\
\hline
Random  & 0.14 & 0.11 & 0.08 \\
DeHateBERT (monolingual) & 0.17 & 0.16 & 0.05 \\
HF German Toxicity Classifier Plus V2 & 0.19 & 0.28 &  \textbf{0.21}\\
HF Multilingual Toxicity Classifier Plus & 0.33 & 0.23 & 0.15\\
Piccolo-HAP (Ours)  & \textbf{0.44} & \textbf{0.34} & \textbf{0.21} \\
\hline
\end{tabular}
\end{adjustbox}
\caption{\label{tab:results-german}
Results on the German TBO dataset . Best results in bold.}
\end{table*}

\subsection{Datasets}

For experiments, we use the following datasets, all of which contain data that is already known to be offensive. The data is converted into a span-selection task, where the classification model is used to identify the toxic spans (and the target of the span when applicable), using the attention maps.

\begin{itemize}[topsep=0pt, itemsep=0pt, leftmargin=.125in, parsep=0pt]
    \item Target Based Offensive Language dataset (TBO) \cite{zampieri-etal-2023-target}: TBO contains around 4500 examples of English twitter data that has been found to be offensive \cite{OLID,SOLID}, providing token-level annotations and identifying both the offensive spans (\textsc{argument}) and its \textsc{target} in the input text. Each tweet can have multiple <\textsc{t,a}> pairs, and may have a "null" target if the target of the offense is not mentioned in the text.  For this demonstration we did not explore the Harmful label assigned to each tweet. We evaluate on the 475 test examples. 
    \item German TBO: We evaluate our model on another language by annotating a small evaluation set of offensive German tweets from the GermEval corpus \cite{germeval}. Two skilled German speaking annotators were trained in the English TBO annotation task, excluding the Harmful label. In total, 255 German tweets were annotated. 
    \item Toxic Spans Detection (TSD) \cite{pavlopoulos-etal-2021-semeval}: The toxic spans detection task (Sem-Eval 2021 Task-5) annotated English toxic comments at the span level, marking spans of text that contribute to the offensive score. They release code that evaluates predictions at the character level.
\end{itemize}

For both TBO datasets, we experiment with using our attention-based approach to identify both the \textsc{target} and \textsc{argument} (\textsc{Target + Arg.}), only the \textsc{argument} (\textsc{Arg. Only}) and only the \textsc{target} (\textsc{Target Only}). In the \textsc{Target Only} setting, we exclude the examples that have no \textsc{target} and only evaluate on the remaining examples; 342 English sentences, and 229 German sentences. We find this to be a fairer evaluation of our attention-based approach, as for sentences without a target the model may still produce argument spans as a prediction (as argument spans will always be attended to heavily). We leave evaluation on the \textsc{null} target examples as future work.

Note, our models are not trained on any of the above datasets, we only use them as a tool to evaluate our attention-based span detection approach for available HAP classification models.

\subsection{Results}

\begin{figure*}[t!]
\centering
\includegraphics[width=\textwidth]{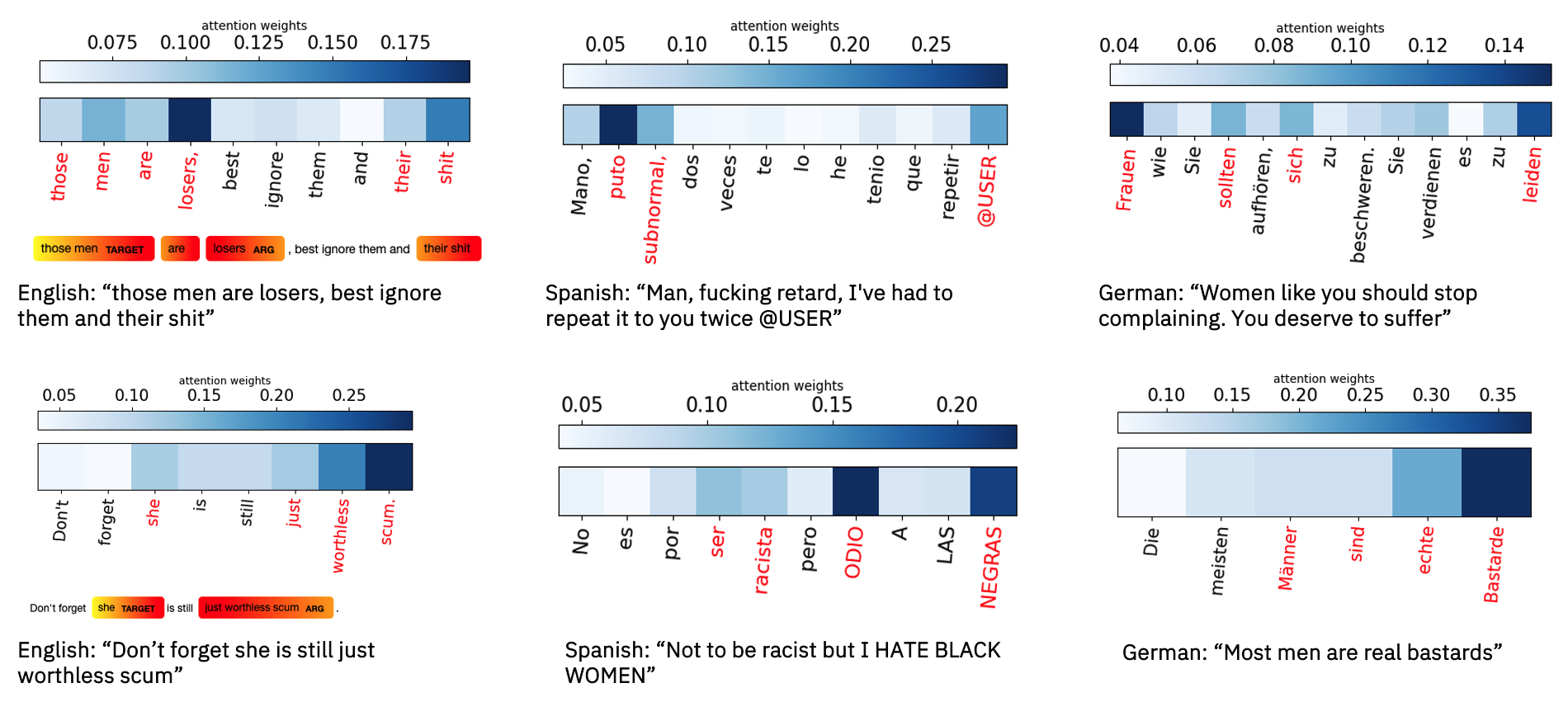}
\caption{System outputs for examples in English, Spanish, and German. Offensive spans and targets marked in red, and images are captioned with the English translations of the input. }
\label{fig-sys-examples}
\end{figure*}  

We re-format each dataset as a span identification task, where the output of our system is the character-level spans for the predicted offensive arguments/targets (the spans are computed using attention maps, as described in \cref{sec:approach}). The F1 scores are computed on a character level, following the approach of \citet{pavlopoulos-etal-2021-semeval}. Here, the training set is used to identify the best attention thresholds to choose the offensive spans, and the test set for evaluating model performance.


\cref{tab:results-english} compares the results of our models to the baseline models on the English datasets. As shown, our model strongly outperforms the HateBERT classification models, the MTC+ Classifier, and the random baseline on the TSD task. We evaluate models on the the TBO dataset under three settings, and show that our models significantly outperform all baselines on identifying both the target and argument, and only the argument. On identifying only the target, it is slightly behind HateBERT finetuned on OffensEval. 

The results on the German TBO dataset are shown in \cref{tab:results-german}. We follow the same experimental setup as for the English results
in \cref{tab:results-english} and present separate results for predicting both target and argument individually and jointly. Our Piccolo-HAP model
outperforms all other German Huggingface models and also the multi-lingual model with the exception of the target-only score by the HF German Toxicity Classifier.

As seen for all models, predicting both \textsc{target} and \textsc{argument} is an easier task than predicting each individually, with Target-only being the hardest setting. A way to improve the performance on this task is to modify the existing method of using the CLS token's attention to identify targets, and instead use the attention of the argument to identify the target. We leave this as future work. 

For further understanding, we analyze a set of sentences from the English TBO dataset for which our model performs poorly in the Target-only setting. We find no clear patterns in this data, however we do find that our approach works very well when the targets themselves are described using offensive or derogatory terms (e.g. "these bitches", "little twats", "clowns", "idiots"). Moreover, our model does not correctly identify targets containing typos (which are common in tweets), such as \textit{yal} instead of \textit{y'all}. As part of future work, a spelling corrector and parser can be built into the HAP prediction system, along with current attention-based thresholds. 
We also analyze our model's output for some test cases where there is a \textsc{null} target annotated in the gold data, and find that our model may predict spans that could be interpreted as the \textsc{target}. For example, the text \textit{"The rich white people don't give a fuck about you unless you affect their bottom line"} marks \textsc{null} target in the gold data, but our model outputs \textit{"the rich white people"} as one span, which could be interpreted as the \textsc{target} of offensive \textsc{argument} (\textit{"don't give a fuck"}). 


\section{System Demonstration} 
We have Jupyter Notebooks and a front-end UI where users can load their models, and obtain visualizations for inputs in any language.

\subsection{Jupyter Notebook}
We have created a Python Jupyter notebook for displaying the <\textsc{t,a}> offensive pairs in a sentence. The notebook will load any encoder-only sentence level offensive classifier. It can be used on multilingual models trained on any language (e.g. English and German as we presented in our experiments). Given a sentence, we generate a heat map using the attention of the model. Then, we identify the offensive \textsc{targets} and \textsc{arguments} using a threshold on the attention. We use the \textit{subj} and \textit{obj} labels from the spaCy dependency parser to identify the \textsc{target} (subject) and \textsc{argument} (object) of offense. Finally, we use the spaCy visualization tools to render the sentence with the offensive \textsc{targets} and \textsc{arguments}\footnote{In English only as the spaCy German parser did not provide the proper information to identify the target and argument.}. Example visualizations for inputs in several languages are shown in \cref{fig-sys-examples}. We would like to extend the tool to more languages based on multi-lingual parsing models.
\subsection{User Interface}

\begin{figure*}[t!]
\centering
\includegraphics[width=0.7\textwidth,height=3.15in]{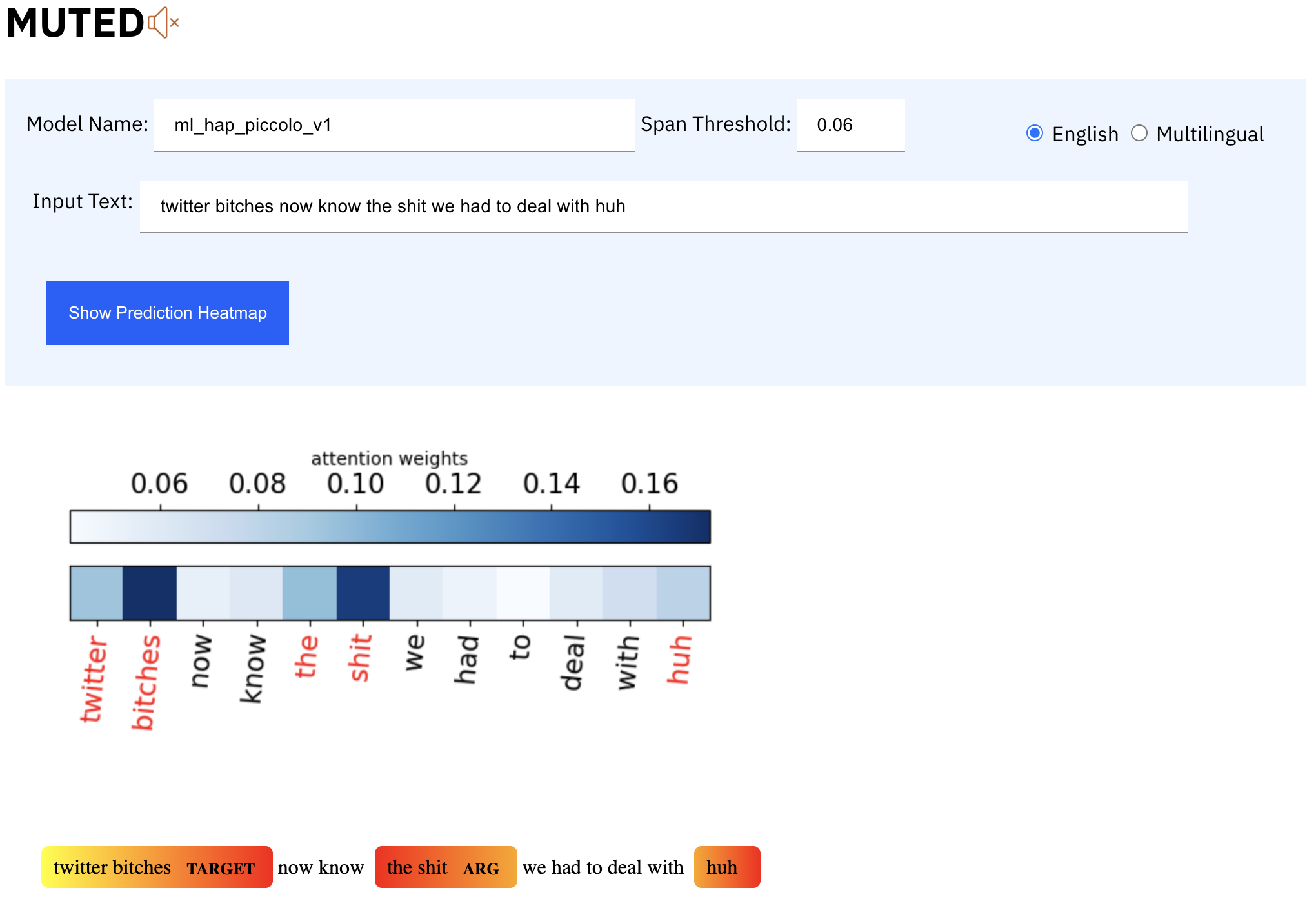}
\caption{Screen-shot of \textsc{Muted} User Interface: The user inputs the model name and input text, and selects the language and attention threshold. The system produces the attention heatmap, and (for English inputs) the spaCy visualization marking the target and argument. }
\label{fig-ui}
\end{figure*}  

The \textsc{Muted} user interface allows the user to play with the HAP classification model without having to know any technical details. The user interface is implemented in Flask which is a lightweight native python web application framework. We show an example of the user interface in \cref{fig-ui}.
The UI allows the user to input the sentence, select if the language is English or non-English and select the value of Span Threshold. Upon clicking "Show prediction Heatmap", the UI renders the output visualizations on the same page. Same page rendering allows the user to tune the output with the best possible parameter values. 

\subsection{System Efficiency}

We evaluate the time taken to produce the predictions and visualizations for a single input by averaging the inference time for 100 English texts. Note that the major difference between the CPU and GPU latencies is contributed by the time taken to make a prediction (which happens on the GPU when available). The visualizations always happen on the CPU, and also utilize more time.

We show the results for two multilingual models- our Piccolo HAP classifier (a 4-layer model with 153 million parameters), and the MTC+ Classifier (a 12-layer model with 277 million parameters). It takes  0.65/0.64s on CPU/GPU to run with our small model, and 0.76/0.65s on CPU/GPU for the base size model, for a single input. \cref{tab:latency} shows the average latency of a single input for the different steps in the process. Thus, the system is quite efficient, and can process 100 examples in about a minute on both CPU and GPU.

\begin{table}[t!]
\centering
\begin{adjustbox}{max width=0.49\textwidth}
\begin{tabular}{|c|c|cc|}
\hline
 & &\textbf{Piccolo} & \textbf{MTC+} \\
 & & \textbf{Model} & \textbf{Model} \\
\hline
\multirow{3}{*}{\textbf{CPU}} & Span Prediction & 0.02 & 0.11 \\
& Attention Map &0.22 &0.23 \\
& SpaCy Visuals & 0.41 & 0.42\\
\hline
\multirow{3}{*}{\textbf{GPU}} & Span Prediction &0.01 & 0.02 \\
& Attention Map &0.22 &0.22 \\
& SpaCy Visuals & 0.41 & 0.41  \\
\hline

\end{tabular}
\end{adjustbox}
\caption{\label{tab:latency}
Time taken (\textit{s}) for span prediction and visualization of a single input. Avg. metric reported over 100 sentences, using a single core CPU and V100 GPU.}
\end{table}

\section{Conclusion}

We present a method for identifying and visualizing offensive arguments and their targets using the attention of the sentence-based offensive classifier to create a heat map. Our multilingual model outperforms existing popular approaches on multiple datasets in English and German. We provide a notebook and user interface to run any multilingual transformer classifier on sentences and visualize the heat map as well as the <\textsc{t,a}> pair using spaCy visualization. In the future, we would like to add a classifier to indicate harm of the <\textsc{t,a}> pair as described in the TBO paper. 
We would also like to extend our demo to provide warnings and hide the offensive content to users.


\section*{Ethics Statement}

\subsection*{Limitations}

In this work, we focus primarily on English and German offensive texts. While our Piccolo model supports 6 languages, and there exists open source HAP classification models of many languages, there is a limitation on datasets available for testing. Specifically, we primarily test on datasets that include annotations of the target of offense, which are not widely available. Creating such datasets for multiple languages would be an interesting direction for future research. Moreover, the test sets that we evaluate on are relatively small in size, and consist of shorter text spans such as tweets.

As mentioned, a more robust way to use transformer attentions to identify the target of the offense is to find the words most heavily attended to by the tokens in offensive span (argument), instead of the CLS vector. In this approach, some type of aggregation strategy would be needed to select the correct tokens from the span, and we aim to implement this as part of future work. 

\subsection*{Intended Use}
Detecting offensive content is an important task that is necessary for avoiding harm. While hateful and harmful content is used to train the models, our intended use is solely for the purpose of avoiding and removing such content and we do not support any malicious or unintended use. 

\subsection*{Biases}
Due to the subjective nature of the task, our German annotated dataset may have unintended biases. These kind of biases are unintentional and will be prevalent in any subjective task. Anyone that uses the data should be aware that such biases may exist. Our TBO annotations are built on top of the existing GermEval dataset \cite{germeval}. We also use the TBO \cite{zampieri-etal-2023-target} and TSD\cite{pavlopoulos-etal-2021-semeval} dataset. Any biases in those original datasets will exist in ours as well which may impact the trained model.  

\section*{Acknowledgement}

We would like to acknowledge the work of our annotators who worked on the German TBO set:
Eva-Maria Wolfe, Joekie Gurski, and Mohamed Nasr.

\bibliography{custom}
\bibliographystyle{acl_natbib}

\appendix



\end{document}